# Explainable AI for Securing Healthcare in IoT-Integrated 6G Wireless Networks

Navneet Kaur
*Dept. of Computer Science*
*University of Missouri-St. Louis*
St. Louis, MO, USA
nk62v@umsystem.edu

Lav Gupta
*Dept. of Computer Science*
*University of Missouri-St. Louis*
St. Louis, MO, USA
lgyn6@umsystem.edu

*Abstract*— As healthcare systems increasingly rely on advanced wireless networks and connected devices, ensuring the security of medical applications has become a critical concern. The integration of Internet of Medical Things (IoMT) devices with real-time health monitoring and care delivery has revolutionized patient care but has also introduced new security vulnerabilities. Each connected device, whether it is a part of a robotic surgical arm, intensive care equipment, or a wearable health monitor, serves as a potential entry point for cyberattacks. Such vulnerabilities could lead to life threatening consequences like poorly performed surgeries, malfunctioning of life support systems or incorrect treatment due to data breaches. The ITU IMT-2030 framework envisions that 6G will be transforming healthcare through massive connectivity, AI, and cloud integration. However, it may also introduce new security vulnerabilities that can threaten the patient safety and privacy. Therefore, addressing these threats requires a thorough reassessment of security measures. This paper presents an innovative use of explainable AI (XAI) techniques – such as SHAP, LIME, and DiCE - to identify vulnerabilities, strengthen security measures, and enhance both security and transparency within the 6G healthcare ecosystem, ensuring robust protection and trust. In addition to the theoretical background, this paper presents experimental analysis and the authors very positive findings.

*Keywords*— Explainable AI, SHAP, LIME, DiCE, Counterfactual, 6G Networks, Healthcare Security, IoMT.

## I. INTRODUCTION

Advancements in wireless communication have greatly enhanced healthcare through innovations like surgical robotics, real-time remote monitoring, and XR-assisted operations in rural and emergency settings [1][2]. The Internet of Medical Things (IoMT) now enables continuous patient data access, facilitating timely interventions, personalized care, and improved patient outcomes [3]. These advancements have set the stage for even greater transformations with the arrival of 6G technology.

The next sixth generation (6G) of wireless communication, is poised to redefine healthcare on an unprecedented scale. Building on the ITU-R M.2160 framework, 6G will introduce AI-native networks, multi-access edge computing (MEC), integrated sensing and communication, and enhanced extended reality (XR) capabilities, unlocking new possibilities for patient care [4]. This framework also envisions immersive XR and haptic sensors to advance remote surgeries and precision care. AI-driven predictive analytics through IoMT devices, such as wearable monitors, will provide real-time health insights, enabling early detection of medical issues, while holographic telepresence could offer supportive patient environments. These context-aware devices powered by advanced sensing and AI will improve interactions between healthcare providers and patients [4][5]. However, these 6G developments bring about significant security challenges. The integration of sensing, edge computing, and cloud services increases threats to connected medical systems, including life-sustaining devices like cardiac monitors and respirators [6]. Vulnerabilities in AI algorithms can lead to model poisoning and erroneous predictions, while edge-based AI is vulnerable to adversarial attacks that could affect critical care decisions [4][7]. Immersive tools like augmented reality in surgeries are prone to unauthorized access, risking incorrect information during critical procedures [8]. Moreover, the fragmented management of edge and cloud infrastructure by multiple operators complicates security and reliability [9].

It is believed that, as the conceptual development of 6G cellular wireless networks progresses, explainable AI (XAI) will become an essential tool for enhancing security, by providing transparent and actionable insights into AI-driven decision-making processes. [10]. This transparency enables healthcare providers to effectively identify and mitigate security threats, facilitating rapid responses to potential breaches [11]. This paper explores the critical role of XAI techniques— Shapley additive explanations (SHAP) [12], Local Interpretable Model-agnostic Explanations (LIME) [13], and Diverse Counterfactual Explanations (DiCE) for IoMT and 6G networks. The novelty of this work lies in applying XAI techniques to WUSTL-HDRL-2024 dataset, specifically designed for medical applications [5] and performing cross-validation of each XAI methods to ensure the consistency and reliability of the explanations. Additionally, the explanations are presented clearly, offering both theoretical insights and practical, actionable strategies for securing medical devices in the evolving 6G networks. The contributions of this paper are summarized as follows:

- Using the ITU-R IMT-2030 framework, we document the security challenges of 6G advancements, focusing on healthcare-related usage scenarios.
- We apply SMOTE balancing to eliminate bias and evaluate the predictive models ability to distinguish normal from anomalous behavior.

- We apply selected XAI techniques - SHAP, LIME, and DiCE – to identify key features that are driving model predictions, revealing potential vulnerabilities in 6G-enabled IoMT environments.
- We perform cross-validation of the results obtained from different XAI methods to evaluate its consistency and reliability, ensuring meaningful explanations of model's prediction.

The following is the structure of the rest of the paper: Section II provides a review of the relevant literature. An extensive discussion of the challenges is provided in Section III. The outcomes and effectiveness of the XAI as an effective solution strategy is shown and discussed in Section IV. Section V offers a thorough validation of the XAI results, while Section VI provides the conclusions.

## II. RELATED WORK AND NOVEL CONTRIBUTION

In recent years, several studies have explored the integration of XAI with healthcare, focusing on enhancing data privacy, interpretability, and clinical decision-making. The authors in [6] propose a healthcare framework that combines explainable AI (XAI) and mass surveillance to improve epidemic monitoring. It utilizes deep learning and edge computing to process health data while ensuring privacy through blockchain technology. In [7] the authors explore the significance of XAI in heart disease diagnosis, emphasizing the need for transparent AI models to foster trust among medical practitioners. The authors discuss interpretability techniques like LIME and SHAP, which clarify the impact of clinical features on predictions. Similarly, in paper [8], the authors introduce a framework combining federated learning (FL) and explainable AI (XAI) to enhance data privacy and model interpretability in next-generation networks. This approach improves prediction accuracy and user trust through interpretable AI decisions. In paper [9], the authors explore machine learning for predicting unplanned hospital readmissions in frail elderly patients using a dataset of over 145,000 records. To improve interpretability, they utilized SHAP, helping clinicians understand the factors behind predictions. Lastly, paper [15] presents a framework for ECG classification using federated learning to improve data privacy and model robustness. It also includes an explainable AI (XAI) component, enabling healthcare professionals to interpret and validate model predictions.

While many existing studies emphasize the importance of AI across various healthcare domains, they often fail to consider the extensive integration of wireless networks and, in the contemporary milieu, the complexity of security challenges introduced by the impending transition from 5G to 6G. Our research attempts to fill this gap. It specifically focuses on the security challenges associated with medical applications in the context of the evolving 6G networks. Unlike prior studies, our study combines more than one XAI method (SHAP, LIME, and DiCE) to provide more comprehensive explanations. These explanations are presented in an accessible manner, making them understandable to non-technical stakeholders or individuals without prior XAI knowledge. Additionally, we cross-validate the outcomes of all XAI methods to confidently assert the correctness of the explanation, thereby fostering greater trust in AI-powered security solutions.

## III. SECURITY CHALLENGES IN 6G IoMT

This section outlines the security challenges associated with the usage scenarios defined in the ITU-R IMT-2030 framework, as presented in document M.2160 [4]. As illustrated in Fig. 1, some of these 6G usage scenarios are extensions of the 5G framework while others, such as., ubiquitous connectivity, AI and communication and integrated sensing and communication, are entirely new scenarios. For a more comprehensive understanding, readers are encouraged to consult the referenced document

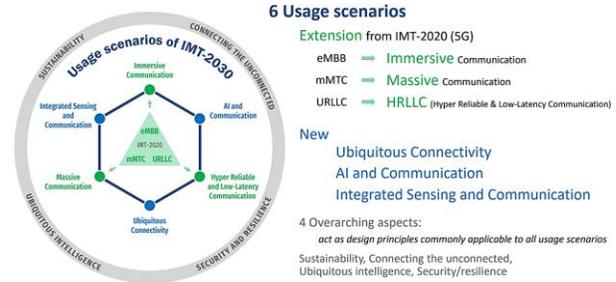

Fig1. Usage Scenarios Wheel Diagram (Source ITU)

### A. 6G Usage Scenarios and their Security Implications

The inherent complexity of the healthcare sector, combined with the diversity and criticality of use cases, underscores the need to address the associated security challenges. In this section, we explore the ITU-defined 6G usage scenarios that impact the healthcare sector and the vulnerabilities they may present

*1) Immersive Communication:* Expanding on 5G's Enhanced Mobile Broadband (eMBB) capabilities, 6G will focus on delivering immersive experiences through enhanced connectivity, ultra-high-definition visuals, spatial audio, and real-time interactivity [4][10]. While 5G improved bandwidth and latency, it struggled with large-scale integration of augmented reality (AR) and virtual reality (VR) [4]. 6G will enable applications like high definition, multi-media remote consultations, surgical simulations, remote surgeries, and collaborative training. For example, surgeons could use AR overlays to access real-time patient data, 3D organ reconstructions, and procedural guides, enhancing precision during complex procedures.

*Security implications:* The expansion of immersive communication could help attackers intercept and potentially alter critical patient information such as heart rate or oxygen saturation levels, leading to life threatening surgical errors. Additionally, the connectivity of immersive tools is susceptible to targeted denial-of-service (DoS) attacks or latency manipulation, which could disrupt the services and hinder a specialist's ability to make accurate assessments in critical situations [11].

*2) Hyper-reliable and low latency communication(HRLLC):* 5G URLLC enables specialized applications requiring ultra-low latency and high reliability, but it still faces challenges like reliability issues and latency spikes that can disrupt time-sensitive medical operations [4]. In 6G, HRLLC will overcome these limitations, ensuring seamless connectivity for critical applications such as remote surgeries and emergency response [11].

*Security implications:* HRLLC presents unique security risks, such as attackers exploiting low-latency demands by launching latency attacks, causing delays or packet losses that lead to time-consuming retransmissions and undetected disruptions in communication channels [16]. Interruptions in complex procedures, like remote robotic surgeries, could reduce precision and put patients at risk, while delays in transmitting crucial biomarkers during emergency responses could undermine the effectiveness of critical medical interventions [17].

*3) Massive Communication:* This scenario enhances the mMTC capabilities of IMT-2020, enabling connectivity for a vast array of IoT applications [4]. While 5G supports a limited network of healthcare devices, 6G will facilitate a significantly larger number of devices to communicate continuously and autonomously, greatly expanding the scope and efficiency of healthcare applications [16].

*Security implications*: An unsecured device, such as a bedside health monitor, could enable hackers to access the entire network, breach sensitive records, manipulate data, or disrupt clinical procedures. For example, a compromised IV pump could allow remote dosage changes, posing serious health risks [17]. In 6G networks, such attacks could spread rapidly, complicating detection and increasing security management challenges.

*4) Ubiquitous Connectivity:* This scenario introduces enhanced coverage and mobility through advanced terrestrial and non-terrestrial networking [4], beyond what is achievable in 5G. In 6G, it aims to provide constant access across devices and locations, devices such as wearable monitors, ambulance IoMT systems, and remote surgical tools, will continuously share data, regardless of their location [16]. For example, during remote surgery, augmented reality (AR) tools will exchange real-time images and surgical inputs with robot-assisted devices to improve patient outcomes [17].

*Security implications:* Universal connectivity can allow any breach to spread rapidly, jeopardizing patient safety by disrupting surgeries or altering critical commands [2].

*5) AI and Communication:* This usage scenario supports distributed computing and AI-powered applications. IMT-2030 will be designed by learning from wireless big data which has yet to be comprehensively exploited. In 6G, AI will improve real-time data analysis, decision-making, and optimization, boosting network efficiency, reducing latency, and enabling intelligent resource management [4].

*Security implications:* AI systems processing real-time health data are susceptible to model or data poisoning attacks, making them vulnerable to cyber threats [2]. Additionally, adversarial attacks can disrupt diagnostic systems, causing incorrect diagnoses and affecting patient care, while compromised predictive analytics may lead to wrong-site surgeries or incorrect patient discharges [4] [9].

*6) Integrated Sensing and Communication:* This usage scenario enables new applications through wide-area, multi-dimensional sensing of IMT-2030 that provides spatial data on both unconnected objects and connected devices, along with their movements and surroundings [4]. In healthcare, this could involve sensors monitoring vital signs, glucose levels, or other health indicators. The data collected is communicated to healthcare providers for immediate analysis and action [2].

*Security implications:* In smart hospitals, sensors that monitor patient movements could be compromised by hackers, potentially disabling distress alarms and putting lives at risk [2]. In remote surgeries, security breaches could inject false data or interfere with surgical instruments, resulting in significant harm. Moreover, tampered location data during emergencies could mislead ambulances, delaying essential care and endangering patients [17].

### B. Mitigation of security risks of 6G use cases

The threats posed by 6G innovations highlight the urgent need for robust protection for patients and the healthcare system. As healthcare increasingly adopts advanced technologies, including AI, these innovations expose vulnerabilities that could be exploited by cyber threats. While AI can help mitigate these threats, its black-box nature often limits transparency and interpretability [18]. XAI emerges as a crucial tool for enhancing transparency and interpretability of AI-driven decisions, empowering security providers to implement stronger protections and maintain a secure healthcare environment amidst evolving cyber threats [19]. By examining how changes in specific features affect model predictions, we can make informed adjustments that improve the stability, reliability, and accuracy of AI models [20].

## IV. EXPERIMENTAL SET-UP & RESULTS

### A. Dataset Description

The WUSTL-HDRL-2024 dataset [5] combines the analysis of both network and host features, leveraging the strengths of Network IDS and Host IDS for security research in medical settings. It includes four major threats: Man-in-the-Middle (MiTM), Distributed Denial of Service (DDoS), Ransomware, and Buffer Overflow attacks as seen in Fig. 2. With a total of 77 features across 145,000 samples (132,000 normal and 13,000 attack instances), this dataset provides a robust and detailed overview of threats for precise analysis.

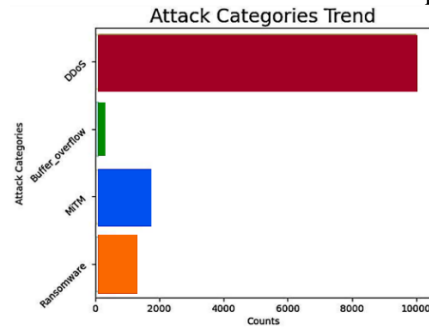

Fig. 2. Visualization Plot of different class types and their counts.

This dataset encompasses critical network infrastructure, including routers, servers, and MEC components, which are integral to 6G environments. It effectively captures the complexities and security challenges that future 6G networks will face, making it an ideal resource for assessing the performance of security measures designed to address emerging threats.

## B. Dataset Pre-Processing

The collected dataset is preprocessed for model development. It is converted into Pandas Data Frames, with infinite and missing values taken care of, duplicates removed, and categorical labels transformed into numerical formats. Columns with missing data were discarded, and normalization is performed. Features with zero variance are excluded, reducing the attribute count from 75 to 52.

To address class imbalances and mitigate potential biases, we employ the Synthetic Minority Over-sampling Technique (SMOTE) [21]. It creates new instances based on the nearest neighbors of the minority class, preserving essential information from the majority class without duplicating the data. Furthermore, we reframed the problem as a binary classification task, distinguishing between attack (1) and normal (0) network traffic, as shown in Fig. 3.

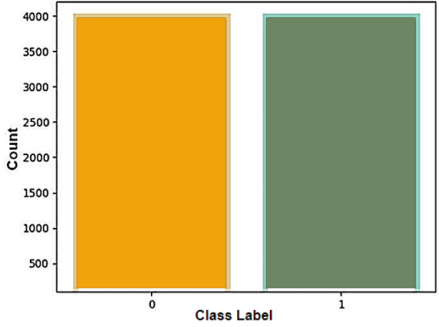

Fig. 3. Distribution of Class Labels – After SMOTE

Given the dataset's size, we randomly selected 2,000 records for each class (normal and attack traffic) to evaluate our machine learning models, as outlined in Table I. Since our primary objective is to develop a model that emphasizes interpretability and prediction clarity while enhancing security in network environments, this dataset size is sufficient to demonstrate the effectiveness of XAI while maintaining computational feasibility.

TABLE I.  DATASET CATEGORIZATION AFTER SMOTE

| Class Types | Total Count (Before SMOTE) | Class Instance (Before SMOTE) | Total Count (After SMOTE) | Class Instance (After SMOTE) |
|---|---|---|---|---|
| Normal | 132884 | 132884 | 2000 | 2000 |
| DDoS | 9971 |  | 500 | 2000 |
| MiTM | 1672 | 12239 | 500 |  |
| Ransomware | 528 |  | 500 |  |
| Buffer_Overflow | 68 |  | 500 |  |

## C. Dataset Modeling

We analyze Logistic Regression, Random Forest, K-Nearest Neighbors, and CNN (with dense layers of 120, 80, 40, and 20 neurons, followed by a single-neuron output layer). Model performance is evaluated using various metrics, the results are summarized in Table II. While further fine-tuning can improve accuracy, our primary focus is ensuring model interpretability through XAI. Given Random Forest's performance, we prioritize it and apply explainable AI techniques to analyze its decision-making and the factors influencing its predictions.

TABLE II.  PERFORMANCE EVALUATION METRIC

| Model | Accuracy | Precision | Recall | F1 Score |
|---|---|---|---|---|
| Logistic Regression | 98.37% | 0.984 | 0.983 | 0.983 |
| CNN | 75.25% | 0.752 | 0.752 | 0.752 |
| Random Forest | 99.85% | 0.998 | 0.998 | 0.998 |
| K-Nearest Neighbor | 98.87% | 0.988 | 0.988 | 0.988 |

## D. Implementation Of Explainable AI Techniques

To explain the decision-making process of the Random Forest model, we selected a random test sample to examine how individual features impact its classification of 'Normal' or 'Attack.' We use different explainable AI methods to enhance the reliability of our findings. By cross-validating results across these techniques, we ensure consistency in the identified features thereby reinforcing confidence in the model's predictions.

### 1) SHAP

SHAP quantifies the contribution of each feature to the model's prediction by assigning specific values to them [12]. The value for feature $i$ in prediction $x$ can be represented as:

$$\emptyset_i(x) = \sum SC\{1,2 \ldots p\}\{i\} \frac{|S|!(p-|S|-1)!}{p!} [f(S \cup \{i\}) - f(S)] \quad (1)$$

Here, $\emptyset_i(x)$ denotes SHAP value for feature $i$ at instance $x$, $p$ denotes the total number of features, $f$ is the model's prediction function, $S$ is the subset of the features excluding $i$, $|S|$ indicates the number of features in subset S, and $|S|!(p-|S|-1)!$ signifies the number of ways to select subset $S$.

Fig 4. illustrates SHAP force plots showing the contribution of each feature to the model's prediction. The base value represents the average model output across the training dataset (approximately 0.4967). Red bars represent high values, while blue bars represent low values. The length of each bar represents the magnitude of each feature's contribution to the model's prediction. Longer bars indicate features that have a stronger influence on the predicted value. The final prediction, displayed at the end of the plot (e.g., 1.00 in this case), represents the cumulative effect of all feature contributions, starting from the base value.

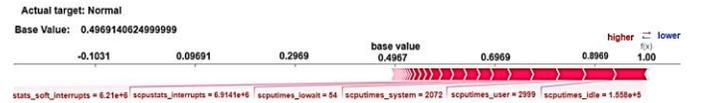

Fig. 4. SHAP Force Plot

We draw the following conclusions from the analysis:
- High value of the soft interrupt ('stats_soft_interrupts') positively impacts the "Normal" prediction, suggesting frequent software interruptions in an active system with ongoing processes and connected devices.
- High system time ('scputimes_system') supports the "Normal" prediction, indicating that the CPU is actively managing core tasks, typical in a normal state handling routine processing, resource management, and device I/O.

- High user time ('scputimes_user') indicates regular CPU usage by applications, suggesting normal operation without overload.
- High idle time ('scputimes_idle') indicates spare CPU capacity, suggesting the system is not stressed and can handle additional tasks smoothly, typical in a "Normal" state.

### 2) LIME (Local Interpretable Model-agnostic Explanations)

LIME approximates the model's prediction function $f$ locally around an instance $x$ [13]. by minimizing a loss function, which can be expressed as:

$$\hat{g} = \arg\min_g L(f, g, \pi^x) + \Omega(g) \quad (2)$$

Here, $\hat{g}$ denotes the interpretable model, $\pi_x$ represents the proximity measure around $x$, L is denoted as loss function, and $\Omega(g)$ is a penalty term ensuring complexity of g.

Fig. 5 illustrates a LIME explanation. In the left panel, the model predicts class [0], indicating a "Normal" status. The middle panel shows each feature's contribution to the prediction. The length of each bar indicates the feature's impact, with longer bars signifying greater influence. The right panel lists all features used for this sample along with their values.

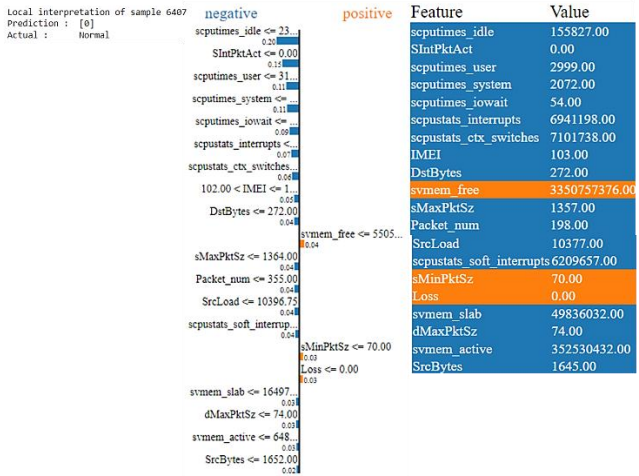

Fig.5. LIME Plot

The following conclusion can be drawn for this test sample:

- The features highlighted in blue contribute negatively to the classification. For instance, excessive idle CPU time ('scputimes_idle') may indicate underutilization, potentially masking malicious processes running in the background. Similarly, unusual network packet activity ('SIntPktAct') could be a sign of a Distributed Denial of Service (DDoS) attack or data exfiltration attempts. Extremely high user CPU time ('scputimes_user') may signal a system overload due to resource-intensive attacks, such as cryptojacking. Additionally, increased I/O wait times ('scputimes_iowait') may indicate system bottlenecks, potentially caused by malicious activities overwhelming system resources. These deviations can shift the model away from predicting "Normal" and suggest a potential attack.
- The orange features like 'svmem_free' (available virtual memory) and 'sMinPktSz' (minimum packet size) contribute positively, reinforcing the prediction of "Normal." They indicate that the system has sufficient resources and normal network behavior, helping balance out the negative contributions of the blue features.

### 3) Diverse Counterfactual Explaination (DiCE)

DiCE aims to generate multiple plausible alternatives (counterfactuals) for a machine learning model's prediction [14]. It can be expressed as:

$$\min_{x'} ||x' - x||^2 + \lambda.\mathcal{L}(f(x', y_{target}) + \beta.Diversity(x', \{x_i\}) \quad (3)$$

Here, $x$ is the original instance, $x'$ is the counterfactual instance, $y_{target}$ is the desired output, $\mathcal{L}$ is a loss function measuring alignment with the target, $||x' - x||^2$ ensures proximity to the original instance and $Diversity(x', \{x_i\})$ promotes variability among counterfactuals.

Fig. 6 illustrates the DiCE explanation. The top row shows the original sample, with a prediction outcome of '1' (indicating "Attack"). The bottom row presents three counterfactual scenarios, with a prediction outcome of '0' (indicating "Normal"). This highlights how minor modifications in the input features can significantly alter the model's prediction outcome. The following conclusion can be drawn from it:

- The features Source Bytes ('SrcBytes') and Destination Bytes ('DstBytes') typically contribute to a "Normal" prediction when their values reflect typical network activity, such as regular data transfers. However, unusually high values could indicate potential abnormal behavior, like data exfiltration or network flooding, pulling the prediction away from "Normal."

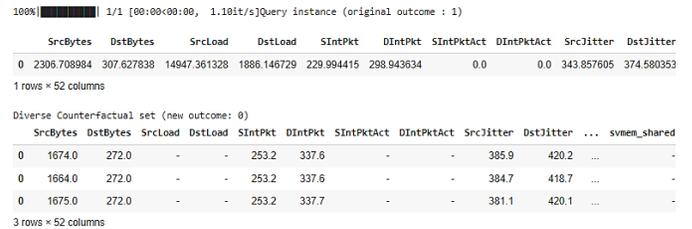

Fig.6. DiCE Explanations

- The features Source Jitter ('SrcJitter') and Destination Jitter ('DstJitter'), which represent variations in packet delay, support a "Normal" prediction when jitter is low, signaling a stable network. High jitter, however, suggests network instability or congestion, which could be a sign of an attack or abnormal network conditions, moving the prediction away from "Normal."

## V. DISCUSSION – VALIDATION OF XAI RESULTS

The interpretability analysis with SHAP, LIME, and DiCE offers critical insights into decision-making, enhancing the model's reliability in distinguishing normal from attack traffic. SHAP associates high system time ('scputimes_system'), user time ('scputimes_user'), and idle time ('scputimes_idle') with normal operations, whereas LIME warns that excessive idle time ('scputimes_idle') may indicate underutilization, potentially masking malicious activity, and high user time ('scputimes_user') could signal cryptojacking or resource-intensive attacks. This suggests that while CPU activity is a strong indicator of normal behavior, extreme deviations can imply anomalies. In network activity, DiCE identifies source bytes ('SrcBytes') and destination bytes ('DstBytes') as indicators of normal traffic, with unusually high values signaling potential data exfiltration or flooding. Similarly, low source jitter ('SrcJitter') and destination jitter ('DstJitter') support a normal prediction, while high jitter suggests network instability or an attack, aligning with LIME's identification of unusual packet activity ('SIntPktAct') as a threat. Regarding system resource availability, LIME highlights available virtual memory ('svmem_free') and minimum packet size ('sMinPktSz') as stabilizing factors in normal predictions, complementing SHAP and DiCE, which focus more on CPU and network metrics. This cross-validation confirms that normal traffic is characterized by stable CPU usage, typical network activity, and sufficient resources, while anomalies in idle time ('scputimes_idle'), jitter ('SrcJitter', 'DstJitter'), and unusual packet behavior ('SIntPktAct') indicate potential attacks. This empirical analysis illustrates how a comprehensive, multi-faceted approach strengthens the security framework of 6G-enabled healthcare systems.

## VI. CONCLUSION

Advancements in wireless networks and healthcare technologies pose significant risks to the integrity of sensitive patient data and the performance of medical applications. As the world prepares for the advent of 6G, we need to find innovative solutions for handling new threats that are likely to manifest due to greater reliance on virtualization, extended reality, and the integration of sensing and AI in communication frameworks. Unchecked, threats arising from an expanded attack surface can result in medical catastrophes and service disruptions. While AI helps address security challenges, its black-box nature limits transparency, making it difficult for security teams to make informed decisions and adjust their security measures effectively. Our research highlights the critical role of Explainable AI (XAI) in overcoming these challenges within 6G networks, specifically for medical applications. By integrating SHAP, LIME, and DiCE, we offer a comprehensive approach to interpreting model decisions, ensuring that security insights are accessible to both technical and non-technical stakeholders. Furthermore, our cross-validation of these XAI methods confirms their reliability and consistency, fostering trust in AI-driven security solutions. Ultimately, the insights from this analysis will guide the development of proactive risk mitigation strategies in the evolving 6G landscape, enhancing the security posture of medical applications.